\documentclass[runningheads]{llncs}
\usepackage[T1]{fontenc}
\usepackage{graphicx}
\usepackage{amsmath,amssymb,amsfonts}
\usepackage[ruled,vlined,linesnumbered]{algorithm2e}
\usepackage{subcaption}
\usepackage{multirow} 
\usepackage{tabularx} 
\usepackage{url} 
\usepackage{array}

\begin{document}
\title{Overlay\_dx - Automating forecasting evaluation}
%
%

\author{Long H. Ngo\inst{1} \and
Mohammed Amine Chamli\inst{1} \and
Jonathan Rivalan\inst{1} \and
Thomas Jaillon\inst{2}}
\authorrunning{Ngo et al.}
%
\institute{Smile, Asni\`eres, France \and
Paris, France\\
}
\maketitle              
\begin{abstract}
\sloppy Traditional evaluation metrics provides numerical values but often lack comprehensibility, hindering effective differentiation of model performances. Our work addresses this challenge by introducing $overlay\_dx$, a novel evaluation metric measuring the performance of time series prediction models. $Overlay\_dx$ is a visual metric that represents the percentage of predictions falling within a confidence interval around actual values. Additionally, once evaluation results are plotted, $overlay\_dx$ computes the area under the overlay curve, providing a quantitative measure of alignment between predicted and actual values across different thresholds and predictions. Through extensive experiments, we demonstrate that our approach offers a unified evaluation framework that combines both visual and numerical assessments, enabling improved model comparison and providing valuable insights for further research and optimization efforts in time series prediction.

\keywords{Machine learning \and Evaluation metric \and Time series data \and Forecasting \and Optimization.}
\end{abstract}
\section{Introduction} \label{intro}

Time series prediction models have become increasingly central in various domains, from financial forecasting to industrial monitoring. However, evaluating these models presents a challenge. Traditional evaluation metrics provide numerical scores that, while mathematically sound, often fail to capture the nuanced performance characteristics that practitioners need to make informed decisions. This limitation becomes particularly apparent when comparing multiple models or when communicating model performance to stakeholders with varying levels of technical expertise.

For example, when comparing two forecasting models using RMSE, a difference between scores of 0.15 and 0.17 provides little intuitive understanding of real-world performance implications. Traditional metrics also fail to capture important aspects like timing of predictions and consistency across different scales, which are crucial for domain-oriented applications such as energy forecasting or financial markets analysis.

Current approaches to time series evaluation typically fall into three categories: point-wise metrics (e.g. Root Mean Squared Error (RMSE), Mean Absolute Error (MAE)), distribution-based metrics (e.g. Kullback–Leibler (KL) divergence), and shape-based metrics (e.g. Dynamic Time Warping (DTW)). While each category offers specific insights, they often fail to provide a unified framework that combines statistical rigor with intuitive interpretability. This fragmentation in evaluation approaches makes it difficult for practitioners to make holistic assessments of model performance.

In this work, we address this fundamental challenge through the introduction of $overlay\_dx$, a novel evaluation metric that combines visual interpretability with quantitative rigor. This new understandable evaluation metric helps to compare the performance of multiple models, not only through its score but also through visualization of prediction performance. The key contributions of our work include: 1) Development of a new visual evaluation metric ($overlay\_dx$) that represents prediction accuracy through an intuitive confidence interval approach; 2) Introduction of a quantitative scoring mechanism based on the area under the overlay curve; 3) Demonstration of the metric's effectiveness across various time series prediction scenarios; 4) Provision of a framework that bridges the gap between technical evaluation and interpretability.

The $overlay\_dx$ metric offers several advantages over traditional evaluation approaches, including enhanced visual interpretability for effective communication with non-technical stakeholders, robustness to outliers and extreme values, ability to capture performance characteristics across varying prediction thresholds, and seamless integration with existing evaluation workflows.

$Overlay\_dx$ is implemented as an open source project to ensure reproducibility and community adoption. The codebase, along with documentation and usage examples, is publicly available on GitHub\footnote{https://github.com/Smile-SA/overlay\_dx}. 
This implementation supports seamless integration with popular machine learning frameworks and includes utilities for visualizing overlay curves and computing Area Under Curve (AUC) scores~\cite{myerson2001area}.

The rest of the paper is organized as follows: Section~\ref{related-work} introduces the related works; Section~\ref{methodology} presents the design principles; Section~\ref{experiments} shows the experiments; Section~\ref{conclusion} concludes the paper while listing the contributions and future works. 

\section{Related works} \label{related-work}

The quality of time series predictions is assessed using evaluation metrics. Evaluation metrics measure the accuracy of predictions by comparing predicted values with actual values. The most commonly used evaluation metrics for assessing predictions include MAE~\cite{willmott2005advantages,hodson2022root} and its family, RMSE~\cite{hodson2022root,chai2014root} and its family, and Akaike's entropy-based Information Criterion (AIC)~\cite{bozdogan1987model,cavanaugh2019akaike}. Below, we review the most commonly used metrics for forecasting models evaluation.

Mean Absolute Error (MAE) represents the average of the absolute differences between predicted and actual values. This measure shows us what level of error to expect in average forecasts. As MAE is an average, it does not identify proportionally very high or low errors. With MAE, lower values indicate better predictions.
\begin{equation}
    \scalebox{0.95}{$MAE = \frac{\sum_{i=1}^{n}\left| Y_{i} - \hat{Y}_{i} \right|}{n}$},
\end{equation}
where $Y_i$ denotes actual values, $\hat{Y}_i$ predicted values, and $n$ the number of predicted values.

Mean Absolute Percentage Error (MAPE)~\cite{de2016mean} represents the proportion of the mean difference between actual and predicted values divided by the actual value. This measure works best with data without zeros and extreme values due to the denominator. The smaller the MAPE, the better the model.
\begin{equation}
    \scalebox{0.95}{$MAPE =\frac{100}{n} \sum_{i=1}^{n}\left| \frac{Y_i - \hat{Y}_i}{Y_i} \right|$}.
\end{equation}

Weighted Mean Absolute Percentage Error (WMAPE)~\cite{cleger2012use} is similar to MAPE, but the errors are weighted according to the absolute value of the target value. This can be useful in preventing large errors in target values from overly influencing the evaluation measure.
\begin{equation}
    \scalebox{0.95}{$WMAPE = \frac{\sum_{i=1}^{n}\left| Y_i - \hat{Y}_i \right|}{\sum_{i=1}^{n}\left| Y_i \right|}$}.
\end{equation}

Mean Squared Error (MSE)~\cite{wang2009mean} is defined as the average of the squared errors. This measure integrates variance and bias, and solves the extreme value and zero problems of MAE and MAPE. The smaller the score, the better the prediction.
\begin{equation}
    \scalebox{0.95}{$MSE =\frac{1}{n} \sum_{i=1}^{n}\left( Y_i - \hat{Y}_i \right)^{2}$}.
\end{equation}

Root Mean Squared Error (RMSE) is defined as the square root of the mean square error (MSE). The RMSE value is in the same unit as the projected value, and we aim to minimize it.
\begin{equation}
    \scalebox{0.95}{$RMSE =\sqrt{\frac{1}{n} \sum_{i=1}^{n}\left( Y_i - \hat{Y}_i \right)^{2}}$}.
\end{equation}

Normalized Root Mean Squared Error (NRMSE)~\cite{shcherbakov2013survey} is a version of RMSE normalized by the mean or difference of the extremums of the actual values. NRMSE is used to compare models on several data sets with different scales.
\begin{equation}
    \scalebox{0.95}{$NRMSE = \frac{RMSE}{mean(y)} ~or~ NRMSE = \frac{RMSE}{y_{max} - y_{min}}$},
\end{equation}
where $Y_{max}$/$Y_{min}$ denote maximum/minimum actual values and $Y$  actual values.

Additionally, Dynamic Time Warping (DTW)~\cite{muller2007dynamic} is a traditional metric that measures similarity between temporal sequences by allowing elastic transformation of time series. Unlike point-wise metrics, DTW can capture phase shifts and temporal distortions. However, its computational complexity and lack of intuitive interpretation limit its practical application.

These approaches, while valuable, often focus on specific aspects of model performance rather than providing a comprehensive evaluation framework. Recent advances in time series evaluation have explored multi-objective metrics that combine multiple aspects of prediction quality. For instance, TIGER~\cite{cummins2011method} incorporates domain-specific constraints into the evaluation framework. However, these approaches often increase complexity without proportionally improving interpretability, highlighting the need for our proposed $overlay\_dx$ method.

While existing metrics have served the field well, they share common limitations: (1) difficulty in interpreting scores in practical terms, (2) sensitivity to outliers and noise, and (3) lack of visual interpretability. These limitations particularly affect practitioners who need to make quick, informed decisions about model selection and optimization. Our proposed $overlay\_dx$ metric directly addresses these gaps while maintaining mathematical rigor.

\section{Methodology} \label{methodology}

The methodology employed in this study involves the implementation of highly visual metrics and measures aimed at enhancing understanding through visualization while exploring new possibilities. Subsequently, a new visual metric is developed and applied to assess the performance of predictive models in time series analysis.

\begin{figure}[b]
    \centering
    \begin{subfigure}[b]{0.55\linewidth}
        \centering
        \includegraphics[width=\linewidth]{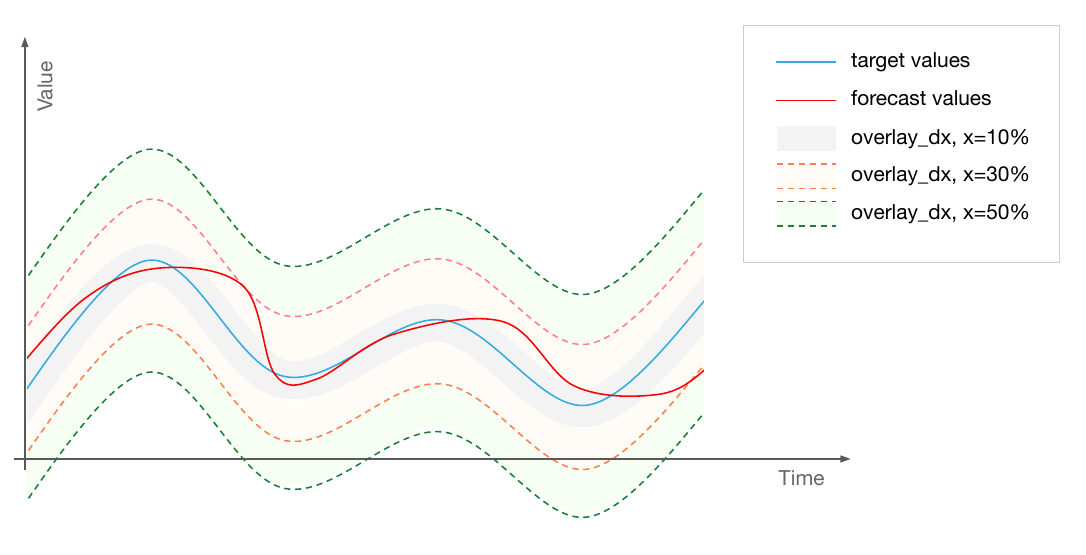}
        \caption{Overlay - overlap rate.}
        \label{fig:overlay-rate}
    \end{subfigure}
    \hfill
    \begin{subfigure}[b]{0.4\linewidth}
        \centering
        \includegraphics[width=\linewidth]{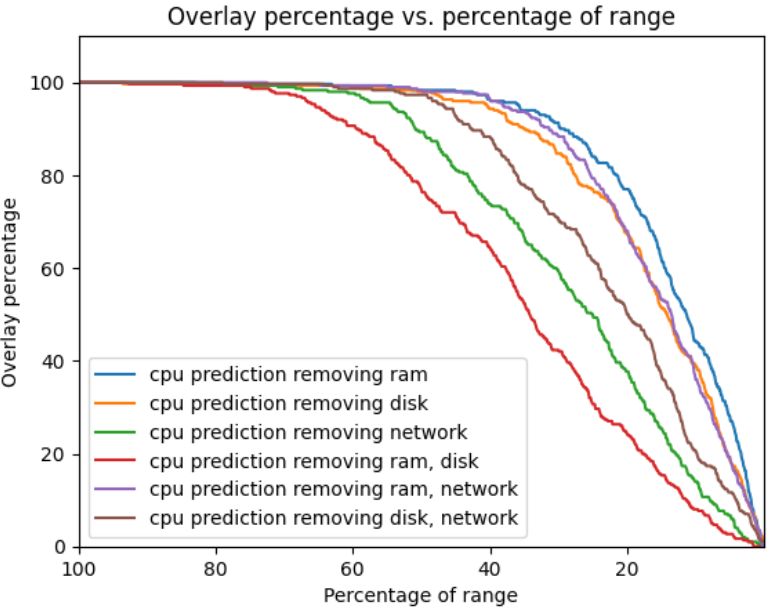}
        \caption{$overlay\_dx$ graphs.}
        \label{fig:overlay-dx}
    \end{subfigure}
    \caption{Overview of $overlay\_dx$ visualisation.}
    \label{fig:overlay-dx-overview}
\end{figure}

For instance, the peak overlap rate (local extrema) was implemented to assess whether predictions were capable of anticipating incidents or extreme values in a time series. Subsequently, the overlay (overlap rate) is defined, representing the percentage of predictions falling within a confidence interval drawn around the actual values of the series. As illustrated in Figure~\ref{fig:overlay-rate}, the overlay indicates the percentage of forecasted values (in red) falling within the intervals (gray, orange, and green) around the target values (in blue). The overlay value varies depending on the size of the interval drawn and can only be equal to or lower than the previous measure when reducing the threshold interval.

By varying the size of the confidence interval, multiple overlay measures were conducted on a single prediction. These interval-based measures can be visualized in a single graph (see Figure~\ref{fig:overlay-dx}, depicting a local CPU usage forecast). In this graph, the interval size (in decreasing order) is plotted on the x-axis, while the overlay measure for that interval is plotted on the y-axis, resulting in the curve profile as depicted in Figure~\ref{fig:overlay-dx}. This curve inevitably decreases as the size of the confidence interval diminishes, allowing for the visual identification of the optimal curve profile, which remains the highest and longest.
Thus, the overlay curve enables the visualization of a model's performance through a curve profile. Consequently, within a single graph, it is possible to present multiple curve profiles and visually compare the performance of several models or configurations. As shown in Figure~\ref{fig:overlay-dx}, it is feasible to quantify and visually measure the performance of one approach compared to another.

Given that the visualization of the $overlay\_dx$ allows for the comparison of several curves, there must also be a numerical metric to differentiate between two very similar curve profiles and make the $overlay\_dx$ an evaluation metric. 
The $overlay\_dx$ score is obtained by evaluating the area under the overlay curve, which represents the cumulative $overlay\_dx$ measures for different thresholds. 
More specifically, $overlay\_dx$ consists of several measures of the overlay metric, which draws an interval around the target values and returns the percentage of forecasted values that fall within this interval. $Overlay\_dx$ calculates different measures of the overlay metric by reducing the size of its interval.
Specifically, $overlay\_dx$ computes the percentage of values where the absolute difference between the forecast and actual values is less than or equal to a threshold $x$. A high score indicates better alignment between predicted and actual values, while a low score indicates a larger deviation from the ideal scenario where perfect alignment is achieved at threshold = 100.
For instance, a score of 77\% represents how well the forecasted values align with the actual values at different thresholds. It indicates that the achieved score is 77\% of the maximum possible score, where perfect alignment would occur at 100\% thresholds.

The score quantifies overall accuracy relative to the ideal scenario. The higher the score, the better the alignment between forecasted and actual values, while a lower score suggests larger deviations. The overlay curve intuitively evaluates model performance, providing insights into accuracy across different thresholds and highlighting areas for optimization.

The major advantage of this new metric lies in its visual representation through the overlay curve.
Unlike traditional numerical measures, the overlay curve allows for a rapid understanding of the model's performance without being significantly influenced by outliers. It provides an overview of the model's accuracy across chosen thresholds.
By examining the overlay curve, it is possible to assess the model's performance for different threshold levels. The point where deviations become more significant can be identified, highlighting areas requiring potential improvement or specific attention.
The overlay curve offers a more nuanced evaluation of model accuracy, considering performance at different thresholds and identifying specific thresholds requiring attention. It also provides a clear and understandable visual representation of model performance, facilitating communication and interpretation of results.

In summary, the use of the $overlay\_dx$ metric and overlay curve enables a more comprehensive evaluation of time series prediction accuracy. It provides a relative measure of alignment between predicted and actual values while offering a clear visualization of model performance. This approach offers more precise perspectives for improvement and optimization by highlighting specific thresholds where deviations become more significant and facilitating the identification of areas requiring further research.
Algorithm~\ref{algo:overlay-dx} summarizes the approach to calculate the $overlay\_dx$ metric with the help of Algorithm~\ref{algo:overlay-function}.

\begin{algorithm}[t]	
	\DontPrintSemicolon
	\SetAlgoLined
	\SetKwInOut{Input}{Input}\SetKwInOut{Output}{Output}
	\Input{x, forecast, target}
    if $x ==0: x =1$
	\BlankLine	
	Calculate the absolute difference between the forecast and actual values: $abs\_diff = abs(target - forecast)$
	
	Count the number of values where the absolute difference is less than or equal to $x$: $num\_overlay = (abs\_diff <= x).sum()$

	\BlankLine
	\Output{Percentage of values that overlay: $pct\_overlay = 100 * num\_overlay / len(y)$}
	\caption{Overlay()}
	\label{algo:overlay-function}
\end{algorithm}

\begin{algorithm}[tb]
    \caption{Overlay\_dx Area Under Curve (AUC) Algorithm}
    \label{algo:overlay-dx}
    \DontPrintSemicolon
    \SetAlgoLined
    \SetKwInOut{Input}{Input}
    \SetKwInOut{Output}{Output}

    \Input{target, forecast, max\_percentage, min\_percentage, step}

    \BlankLine
    Compute the value range of the target: 
    $value\_range = \max(target) - \min(target)$
    
    \BlankLine
    Generate a range of percentages: 
    $percentages = \text{np.arange(max\_percentage, min\_percentage, -step)}$
    
    \BlankLine
    Initialize an empty list: 
    $overlay\_percentages = [\ ]$
    
    \BlankLine
    \For{$pct \in percentages$}{
        Compute tolerence $x$: 
        $x = \frac{pct}{100} \cdot \frac{value\_range}{2}$
        
        \BlankLine
        Compute the overlay percentage using Algorithm~\ref{algo:overlay-function}: 
        $overlay\_pct = \text{Overlay}(x, \text{forecast}, \text{target})$
        
        \BlankLine
        Append $overlay\_pct$ to $overlay\_percentages$
    }
    
    \BlankLine
    \Output{$overlay\_dx = AUC(overlay\_percentages)/(max\_percentage\cdot100)$}
\end{algorithm}

\section{Experiments} \label{experiments}
To demonstrate the effectiveness and utility of the $overlay\_dx$ metric, we conducted extensive experiments using both synthetic and real-world time series data. These experiments aimed to compare $overlay\_dx$ with traditional metrics, assess its robustness to outliers, demonstrate its utility in model selection, and validate its applicability across different types of time series data.

\subsection{Experimental Setup}

To validate the performance of $overlay\_dx$, we utilized three real-world datasets, namely Beijing Multi-Site Air-Quality (Pollution)~\cite{zhang2017cautionary}, Electricity Transformer Temperature (ETT)~\cite{haoyietal-informer-2021}, and Electricity Load Diagrams 2011-2014 (ELD)~\cite{electricityloaddiagrams20112014_321} datasets and a simulated dataset.

The simulated dataset comprises diverse time series, each derived from a base curve modified by adding noise, introducing outliers, bias, delay, or other alterations (see Table~\ref{tab:simulated-dataset}). The base curve represents the target, while the modified versions simulate predictions made by a model. The objective is to showcase diverse scenarios and evaluate $overlay\_dx$'s effectiveness across various situations.

\begin{table}[tb]
\centering
\caption{Generated groups of time series and their variations.}
\renewcommand{\arraystretch}{1} 
\small 
\begin{tabularx}{\textwidth}{|m{2.2cm}|>{\centering\arraybackslash}m{3.3cm}|>{\raggedleft\arraybackslash}m{6.35cm}|}
\hline
\textbf{Group Name} & \textbf{Baseline Function} & \textbf{Added Variations} \\
\hline
Constant & $f(x) = 100$ & Noise, bias, delay, outliers \\
\hline
Linear Trend & $f(x) = x$ & Misestimate, lag, bias, step changes \\
\hline
Seasonal & $50 + 30 \cdot \sin(x)$ & Amplitude, frequency, asymmetry errors; missed peaks, phase shift\\
\hline
Random Walk & $\sum_{i=1}^{n} X_i$ - normal random vars & Smoothed, delay, noise, trend-bias, regime shifts \\
\hline
Multiple Seasonality & $50 + 20 \cdot \sin(x) + 10 \cdot \sin(7x)$ & Missed short cycle, amplitude ratio error, noise \\
\hline
Trend Change & $x$ if $x \in [0, 50]$ else $50 - x$ & Missed reversal, late detection, overreaction \\
\hline
Cyclic  & $x + \text{sawtooth}(x)$ & Trend/cycle only, magnitude error\\
\hline
Exponential Growth & $\exp(x)$ & Linear approximation, misestimate, delayed response, noise \\
\hline
\end{tabularx}
\label{tab:simulated-dataset}
\end{table}

The pollution dataset~\cite{zhang2017cautionary} includes hourly air pollutant measurements from 12 air quality monitoring stations in Beijing. It includes data on six major air pollutants along with six related meteorological variables. The data, covering 2013 to 2017, were sourced from the Beijing Municipal Environmental Monitoring Center and local weather stations operated by the China Meteorological Administration. Missing values are represented as NA.

The ETT dataset~\cite{haoyietal-informer-2021} contains two years of data from two counties in China, focusing on long-term electric power system deployment. It includes subsets at the 1-hour ({ETTh1, ETTh2}) and 15-minute (ETTm1) levels, with each data point featuring the target variable "oil temperature" and six power load features. It is split into training, validation, and test sets at a 12/4/4-month ratio.

The ELD dataset~\cite{electricityloaddiagrams20112014_321} contains electricity consumption of 370 points/ clients from 2011 to 2014.

We selected eight time series forecasting methods for comparison on the two real-world datasets, including Linear Regression~\cite{aalen1989linear}, Random Forest Regressor~\cite{segal2004machine,schonlau2020random}, XGBoost Regressor~\cite{zhang2020predicting}, LightGBM Regressor~\cite{shehadeh2021machine}, K-nearest Neighbors Regressor~\cite{peterson2009k,song2017efficient}, Extra Trees Regressor~\cite{ahmad2018predictive}, Bagging Regressor~\cite{aslam2023forecasting}, and ARIMA~\cite{ariyo2014stock}.

In our experiments, we used TimeSeriesSplit\footnote{\begingroup \scriptsize https://scikit-learn.org/1.6/modules/generated/sklearn.model\_selection.TimeSeriesSplit.html \endgroup} as the cross-validator. Unlike traditional k-fold cross-validation, TimeSeriesSplit maintains the chronological order of observations. In the \textit{k-th} split, the first \textit{k} folds serve as the training set, and the \textit{(k+1)-th} fold is used for testing, mimicking real-world scenarios where models are trained on historical data to predict future outcomes.

\subsection{Experimental Results}

\begin{figure}[tb]
    \centering
    \includegraphics[width=0.5\linewidth]{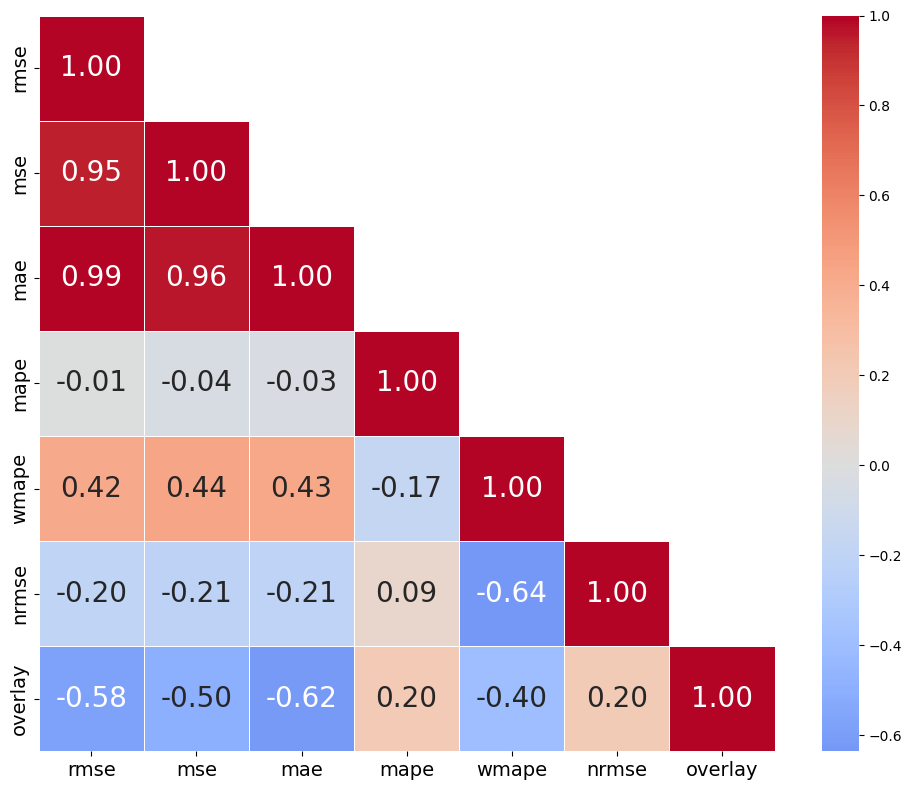}
    \caption{Heat map of metrics correlation on the simulated dataset.}
    \label{fig:correlation-heatmap}
\end{figure}

To ensure robust evaluation, we conducted experiments across multiple dimensions. First, we calculated the correlation of $overlay\_dx$ and other metrics using the simulated dataset. The heat map in Figure~\ref{fig:correlation-heatmap} reveals a high negative correlation between the $overlay\_dx$ and absolute error metrics (MAE, RMSE, MSE), with values ranging from -0.62 to -0.5, implies that $overlay\_dx$ prioritizes absolute improvements in error magnitudes.
It has weak positive correlations with percentage-based metrics (MAPE, NRMSE), suggesting that its interpretation may differ for datasets where relative errors are more critical. The weak correlation with WMAPE suggests $overlay\_dx$ is somewhat effective in capturing weighted errors but not as strongly as absolute error metrics. In conclusion, the varying correlations demonstrate that $overlay\_dx$ captures different performance aspects compared to traditional metrics, providing a complementary perspective on model accuracy.

We then compared $overlay\_dx$ with traditional metrics, including MAE, RMSE, MSE, MAE, MAPE, WMAPE, and NRSME, across all the models on the two real-world datasets. 

\begin{figure}[htb]
    \centering
    \begin{subfigure}[b]{0.3\linewidth}
        \centering
        \includegraphics[width=\linewidth]{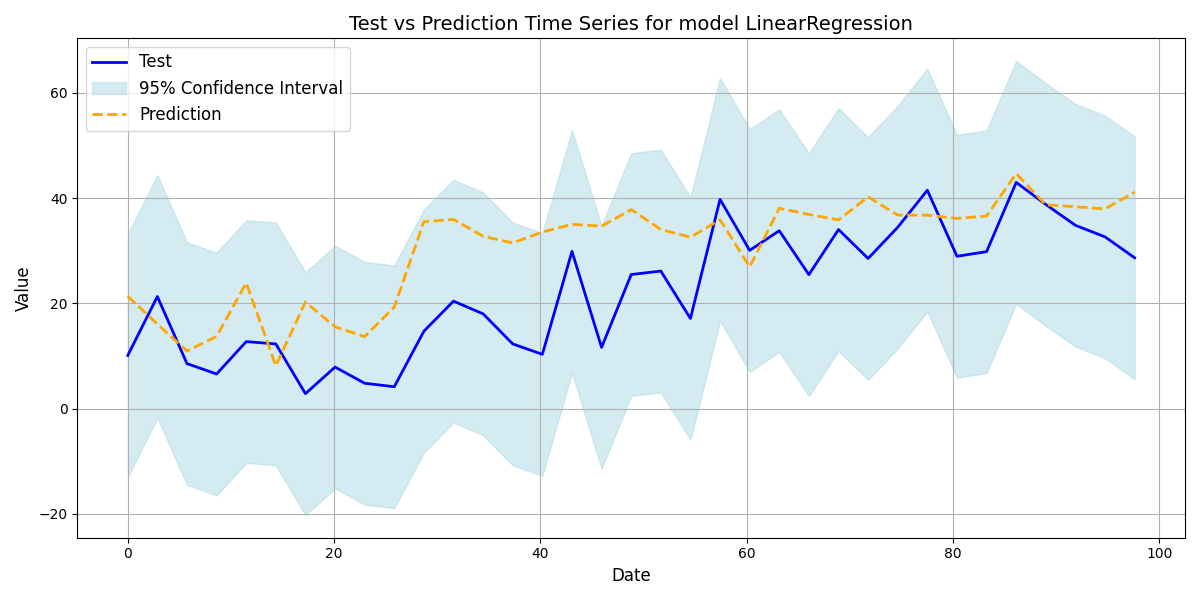}
        \caption{LinearRegression}
    \end{subfigure}
    \hfill
    \begin{subfigure}[b]{0.3\linewidth}
        \centering
        \includegraphics[width=\linewidth]{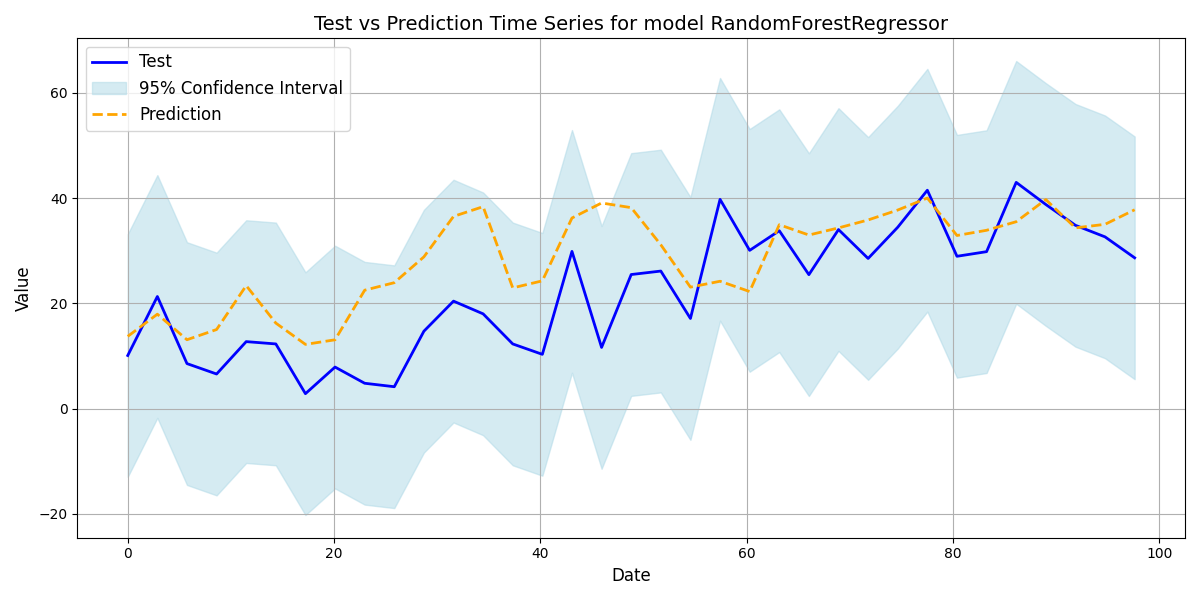}
        \caption{RandomForest}
    \end{subfigure}
    \hfill
    \begin{subfigure}[b]{0.3\linewidth}
        \centering
        \includegraphics[width=\linewidth]{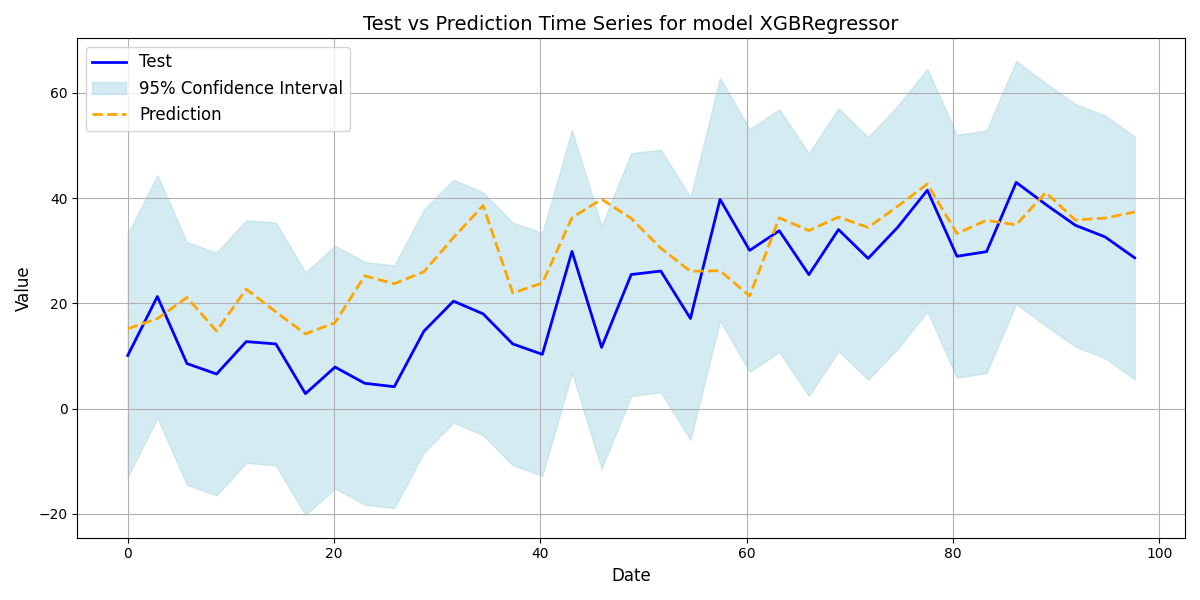}
        \caption{XGBRegressor}
    \end{subfigure}
    \hfill
    \begin{subfigure}[b]{0.3\linewidth}
        \centering
        \includegraphics[width=\linewidth]{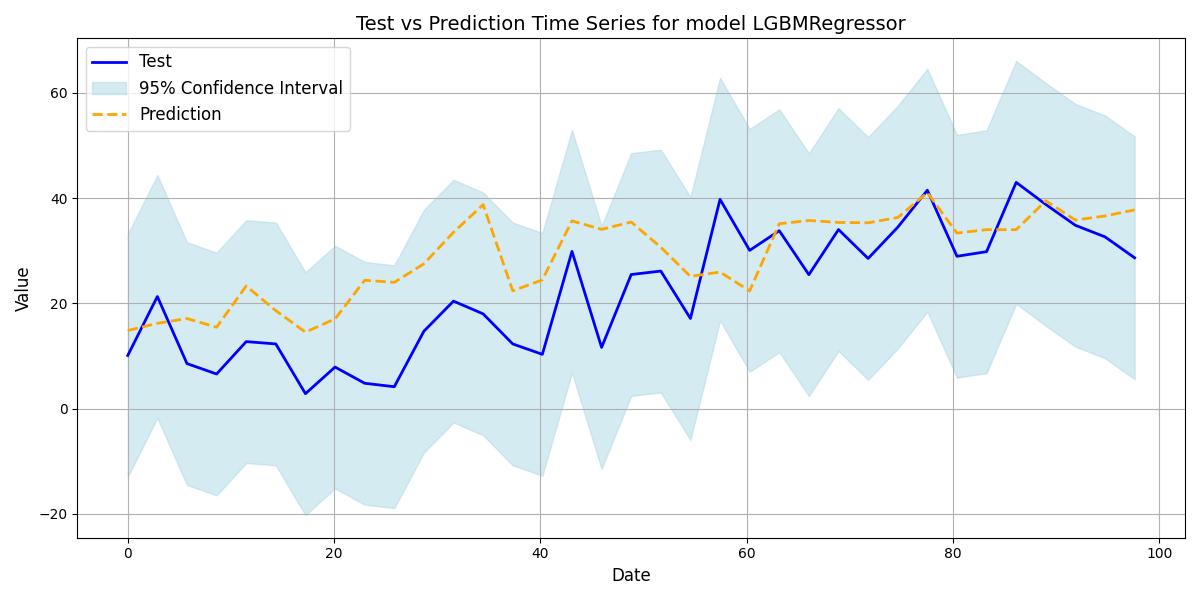}
        \caption{LGBMRegressor}
    \end{subfigure}
    \hfill
    \begin{subfigure}[b]{0.3\linewidth}
        \centering
        \includegraphics[width=\linewidth]{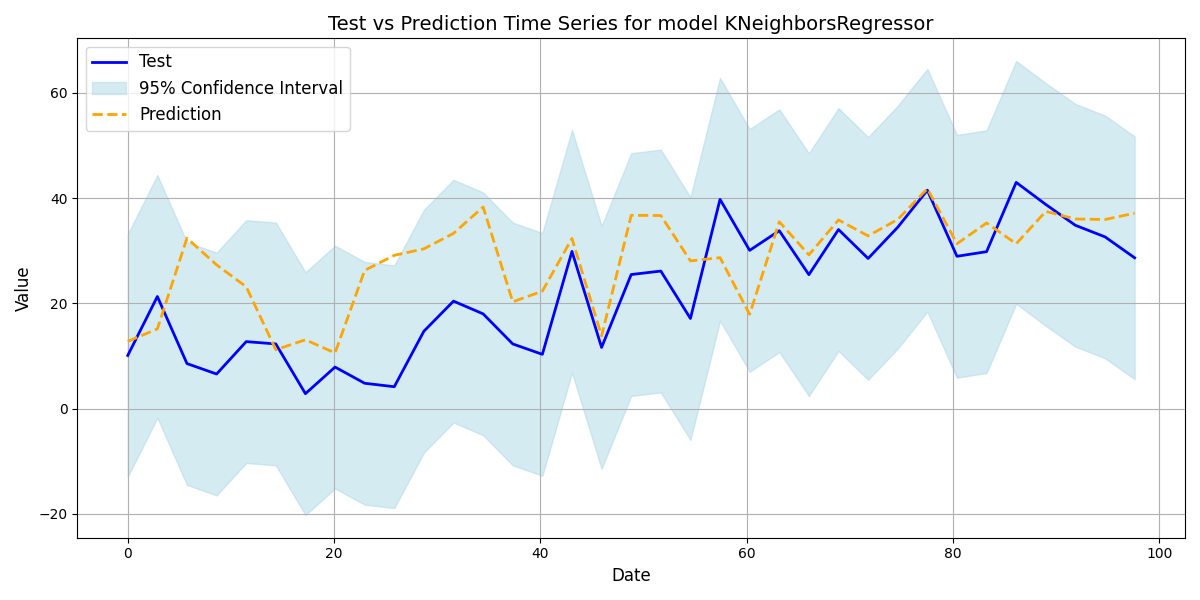}
        \caption{KNeighborsRegressor}
    \end{subfigure}
    \hfill
    \begin{subfigure}[b]{0.3\linewidth}
        \centering
        \includegraphics[width=\linewidth]{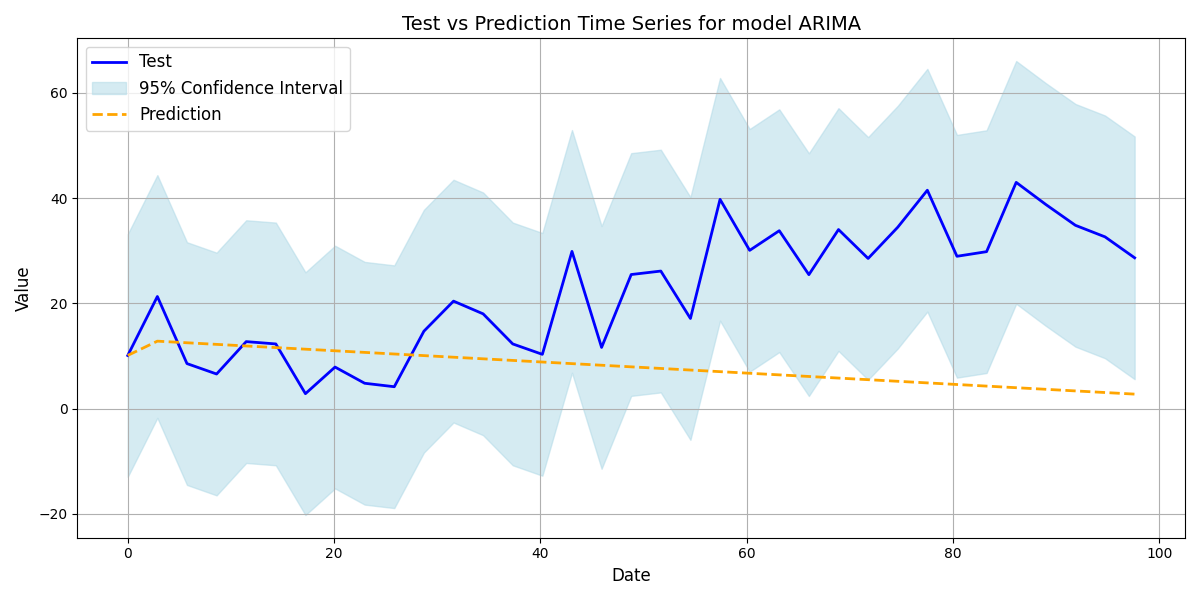}
        \caption{ARIMA}
    \end{subfigure}
    \caption{The predictions of 6 forecasting methods on the ETT dataset. The orange/ blue curves represent predictions/ ground truth values.}
    \label{fig:predicts-ETT}
\end{figure}

Detailed error analysis revealed distinct performance patterns across different time scales. Figure~\ref{fig:predicts-ETT} illustrates the visual prediction results of 6 forecasting methods on the ETT dataset. Visualization of predictions vs. actual values shows that all models sometimes struggle with sudden extreme events. 

\begin{figure}[htb]
    \centering
    \begin{subfigure}[b]{0.48\linewidth}
        \centering
        \includegraphics[width=\linewidth]{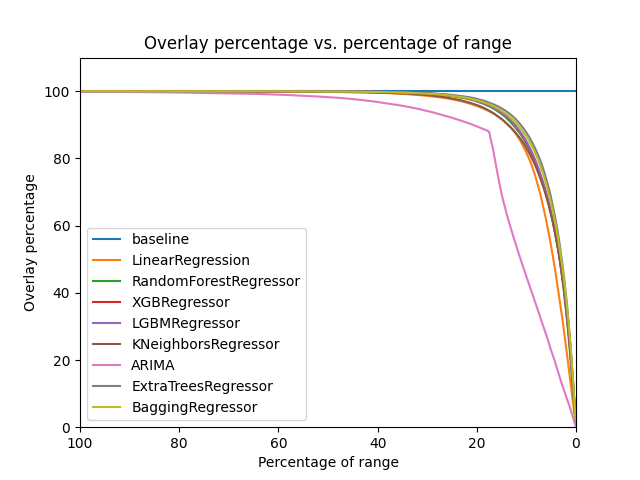}
        \caption{Pollution dataset}
    \end{subfigure}
    \hfill
    \begin{subfigure}[b]{0.48\linewidth}
        \centering
        \includegraphics[width=\linewidth]{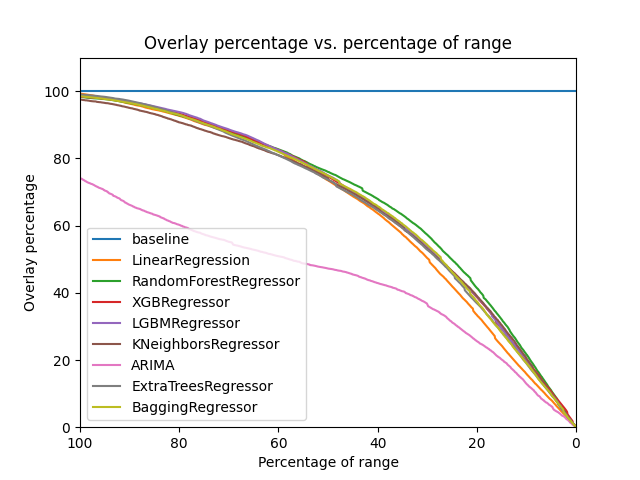}
        \caption{ETT dataset}
    \end{subfigure}
    \hfill
    \begin{subfigure}[b]{0.48\linewidth}
        \centering
        \includegraphics[width=\linewidth]{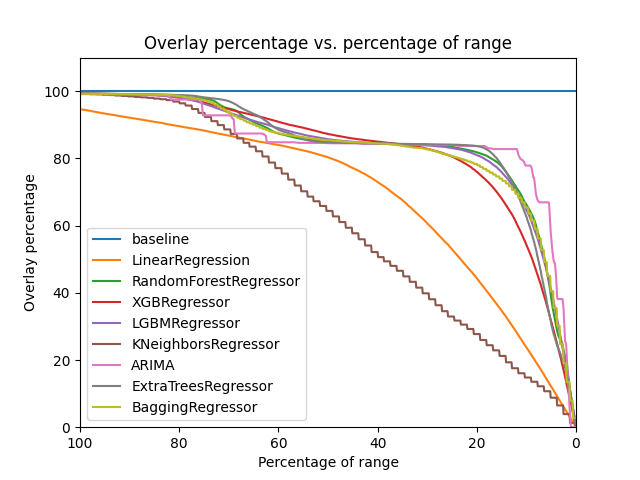}
        \caption{ELD dataset}
    \end{subfigure}
    \caption{$overlay\_dx$ graphs of various methods on 3 datasets.}
    \label{fig:overlay-dx-graph}
\end{figure}

Table~\ref{tab:results}, which presents the $overlay\_dx$ scores and traditional metrics, and Figure~\ref{fig:overlay-dx-graph}, which shows the $overlay\_dx$ graphs for all models, demonstrate how $overlay\_dx$ metric offers a nuanced evaluation framework that complements traditional metrics by capturing performance across multiple confidence intervals. In the pollution dataset, ExtraTreesRegressor demonstrates superior performance with the highest $overlay\_dx$ score of 0.952, accompanied by the lowest RMSE (36.018), MAE (25.9096), and MAPE (0.591). The WMAPE of $0.00$ for ARIMA highlights a computational anomaly, underscoring the limitations of this metric. This suggests that the WMAPE calculation for ARIMA might have encountered a computational edge case problem.  Similarly, in the ETT and ELD datasets, RandomForestRegressor and ARIMA lead with $overlay\_dx$ scores of 0.674 and 0.847, respectively, showcasing the metric's ability to provide comprehensive model assessment beyond single-point evaluations.

\begin{table}[htb]
\centering
\caption{Performance metrics of different forecasting models for three datasets.}
\resizebox{\textwidth}{!}{%
\begin{tabular}{|c|c|c|c|c|c|c|c|c|}
\hline
\textbf{Dataset} & \textbf{Model} & \textbf{rmse} & \textbf{mse} & \textbf{mae} & \textbf{mape} & \textbf{wmape} & \textbf{nrmse} & \textbf{overlay} \\ \hline
\multirow{8}{*}{Pollution} & LinearRegression & 47.410 & 2247.729 & 32.028 & 0.870 & 0.377 & 0.559 & 0.935 \\ \cline{2-9}
 & RandomForestRegressor & 40.325 & 1626.087 & 26.639 & 0.660 & 0.314 & 0.475 & 0.946 \\ \cline{2-9}
 & XGBRegressor & 39.195 & 1536.277 & 26.020 & 0.630 & 0.307 & 0.462 & 0.947 \\ \cline{2-9}
 & LGBMRegressor & 39.110 & 1529.555 & 25.910 & 0.620 & 0.305 & 0.461 & 0.947 \\ \cline{2-9}
 & KNeighborsRegressor & 43.781 & 1916.793 & 27.729 & 0.629 & 0.327 & 0.516 & 0.943 \\ \cline{2-9}
 & ExtraTreesRegressor & \textbf{36.018} & \textbf{1297.328} & \textbf{23.338} & \textbf{0.591} & 0.275 & \textbf{0.424} & \textbf{0.952} \\ \cline{2-9}
 & BaggingRegressor & 38.0439 & 1447.336 & 24.457 & 0.607 & 0.288 & 0.448 & 0.950 \\ \cline{2-9}
 & ARIMA & 84.4496 & 7131.738 & 62.472 & 2.814 & \textbf{0.00} & 0.995 & 0.874 \\ \hline \hline
\multirow{8}{*}{ETT} & LinearRegression & 11.632 & 135.308 & 9.467 & 1.835 & 0.414 & 0.509 & 0.646 \\ \cline{2-9}
 & RandomForestRegressor & \textbf{11.219} & \textbf{125.862} & \textbf{8.723} & 2.347 & \textbf{0.381} & \textbf{0.490} & \textbf{0.674} \\ \cline{2-9}
 & XGBRegressor & 11.441 & 130.907 & 9.045 & 2.22 & 0.395 & 0.500 & 0.663 \\ \cline{2-9}
 & LGBMRegressor & 11.333 & 128.447 & 8.998 & 2.281 & 0.393 & 0.495 & 0.664 \\ \cline{2-9}
 & KNeighborsRegressor & 12.007 & 144.173 & 9.336 & 2.697 & 0.408 & 0.525 & 0.654 \\ \cline{2-9}
 & ExtraTreesRegressor & 11.523 & 132.769 & 9.205 & 2.649 & 0.402 & 0.504 & 0.656 \\ \cline{2-9}
 & BaggingRegressor & 11.508 & 132.424 & 9.113 & 2.568 & 0.398 & 0.503 & 0.660 \\ \cline{2-9}
 & ARIMA & 20.853 & 434.866 & 16.784 & \textbf{1.531} & 0.734 & 0.912 & 0.435 \\ \hline
 \hline
 \multirow{8}{*}{ELD} & LinearRegression & 8.875 & 78.766 & 6.423 & 2.2E15 & 1.583 & 2.187 & 0.683 \\ \cline{2-9}
 & RandomForestRegressor & 5.786 & 33.482 & 3.222 & 1.4E15 & 0.794 & 1.426 & 0.838 \\ \cline{2-9}
 & XGBRegressor & 5.747 & 33.024 & 3.514 & 1.5E15 & 0.866 & 1.416 & 0.823 \\ \cline{2-9}
 & LGBMRegressor & 5.813 & 33.791 & 3.256 & 1.3E15 & 0.803 & 1.433 & 0.837 \\ \cline{2-9}
 & KNeighborsRegressor & 9.101 & 82.820 & 7.688 & 5.1E15 & 1.895 & 2.243 & 0.610 \\ \cline{2-9}
 
 & ExtraTreesRegressor & \textbf{5.632} & \textbf{31.718} & 3.242 & 1.4E15 & 0.800 & \textbf{1.388} & 0.837 \\ \cline{2-9}
 & BaggingRegressor & 5.916 & 34.995 & 3.358 & 1.5E15 & 0.828 & 1.458 & 0.831 \\ \cline{2-9}
 & ARIMA & 6.079 & 36.957 & \textbf{3.059} & \textbf{0.8E15} & \textbf{0.754} & 1.498 & \textbf{0.847} \\ \hline 
\end{tabular}%
}
\label{tab:results}
\end{table}

\section{Conclusion and future works} \label{conclusion}

Our introduction of $overlay\_dx$ represents a significant advancement in evaluating time series prediction models by combining visual interpretability and quantitative assessments into a comprehensive framework, addressing a critical gap in existing methods. The overlay curve provides an intuitive visual representation of model performance, allowing for rapid understanding and identification of areas for improvement, while $overlay\_dx$ offers a quantitative measure of alignment between predicted and actual values. This dual approach enhances the evaluation process, making it both interpretable and precise. Additionally, $overlay\_dx$ serves as a comprehensive evaluation framework, accommodating both technical and non-technical stakeholders to facilitate better communication and decision-making in model selection. 
Through extensive experimentation, we have demonstrated its reliability across various datasets, model types, and prediction scenarios. 

While $overlay\_dx$ offers significant advantages, we acknowledge several limitations. While the computational complexity increases with the number of threshold levels evaluated, the choice of threshold ranges can influence the final score. Also, the visual interpretation may still require some training for optimal use.

This work opens several promising leads for future research. One direction involves the development of algorithms for automated threshold selection, enabling the determination of optimal threshold ranges for various types of time series data. Another is the extension of $overlay\_dx$ to accommodate multivariate time series prediction evaluation. Exploring real-time evaluation methods, such as streaming variants of $overlay\_dx$ for online assessment of prediction models, also presents valuable opportunities. Additionally, integrating $overlay\_dx$ into deep learning frameworks by developing $overlay\_dx$-based loss functions for neural network training could enhance model performance. Finally, creating domain-specific adaptations of $overlay\_dx$ for specialized applications, such as financial forecasting or weather prediction, offers significant potential for targeted advancements.

We believe $overlay\_dx$ represents a significant step forward in time series model evaluation, providing a foundation for future research in the forecasting area. The metric's ability to combine visual interpretability with quantitative rigor addresses a fundamental need in the field, and its extensibility provides numerous opportunities for future development and application.

\begin{credits}
\subsubsection{\ackname} This work has been funded by the European Union’s Horizon Europe research and innovation program under grant agreement No. 101070487 (NEPHELE). 
\end{credits}
%
%
%
\bibliographystyle{splncs04}
\bibliography{mybibfile}
\end{document}